\def\year{2021}\relax
\documentclass[letterpaper]{article} % DO NOT CHANGE THIS
\usepackage{aaai21}  % DO NOT CHANGE THIS
\usepackage{times}  % DO NOT CHANGE THIS
\usepackage{helvet} % DO NOT CHANGE THIS
\usepackage{courier}  % DO NOT CHANGE THIS
\usepackage[hyphens]{url}  % DO NOT CHANGE THIS
\usepackage{graphicx} % DO NOT CHANGE THIS
\usepackage{natbib}  % DO NOT CHANGE THIS AND DO NOT ADD ANY OPTIONS TO IT
\usepackage{caption} % DO NOT CHANGE THIS AND DO NOT ADD ANY OPTIONS TO IT
\usepackage{amsmath}

\DeclareMathOperator*{\argmax}{arg\,max}

\usepackage[ruled,vlined]{algorithm2e}
\SetAlFnt{\small}
\SetAlCapNameFnt{\small}
\SetAlCapFnt{\small}
\usepackage{float}
\title{Searching k-Optimal Goals for an Orienteering Problem on a Specialized Graph with Budget Constraints}
\author{

    Abhinav Sharma,
    Advait Deshpande,
    Yanming Wang,
    Xinyi Xu,
    Prashan Madumal,
    Anbin Hou
}
\affiliations{
    University of Melbourne\\
    Parkville Campus\\
    Victoria, Australia 3010\\
}
\begin{document}

\maketitle

\begin{abstract}
We propose a novel non-randomized anytime orienteering algorithm for finding k-optimal goals that maximize reward on a specialized graph with budget constraints. This specialized graph represents a real-world scenario which is analogous to an orienteering problem of finding k-most optimal goal states.
\end{abstract}

\section{Introduction}
% Informative Path Planning (IPP) is an exquisite planning problem that aims at finding the most informed path from a pre-defined start state to a goal state that maximises the reward with respect to the budget constraints. 
Orienteering Problem (OP) is a special case of the Informative Path Planning (IPP) problem where rewards at different nodes are calculated independently of each other. However, the OP is considered to be NP-hard and mostly solved with heuristic-based search strategies and customized algorithms \cite{Wei2020}. We aim to solve a domain-related orienteering problem which can be formalized for a specialized directed weighted graph. First, we initialize a specialized graph for mapping the Parkville campus of the University of Melbourne. We then use this graph to formalize our problem of finding the most optimal nearest building from a starting building such that the reward can be maximized within the provided travelling budget constraint. The proposed non-randomized algorithm is applied to find k-most optimal nearest buildings inside the campus from a given starting building, discussed in the results section. We also show how COVID-19 lock-down restrictions can be incorporated into our algorithm to solve our defined orienteering problem.

\section{Problem Formulation}
We formulate our domain-related optimal building finding problem into a generic orienteering problem (OP) for a specialized graph below.

Let us assume a weighted directed specialized graph $G_s = (V,E)$ for $n$ number of nodes where $v_s \in V$ is the pre-defined start node such that $
    V = \{v_1, v_2, v_3,...,v_n\}$
and $
    E = \{(v_s, v_i)\:\backslash\:(v_s, v_s)\:|\:\forall\:i \in [1,n]\:\}
$. Here, $v_s$ is having $n$ out-degree with 0 in-degree (i.e. $v_s$ is connected to every other node in $V$) and $v_i\:\forall\:i \in [1,n]$ \textbackslash $\:v_s$ is connected to only $v_s$ with 1 in-degree and 0 out-degree. 

Let $v_g$ be the set of $k$-optimal goal nodes s.t. $v_g \in V$ and $k \leq n$. These goals are attained in the decreasing order of their gained rewards after respecting budget constraints (i.e. $v_{g1} > v_{g2} > ... > v_{gk}$). Let $r$ be the set of nodes which we can visit such that $r \subseteq V$ \textbackslash\:$v_s$. Let $B$ be the travelling budget which will enable the budget constraints. Let $O$ be the generic objectives and $F$ be the generic factors which can be used to tweak the reward function of the problem. 

For each $r$, let $R(r, o, f)$ be the reward function where $\:R:(r,o,f) \rightarrow {\rm I\!R}_0^+ \cup \{\infty\}$ calculates the reward based on the provided set of factors $f \subseteq F$ and objective $o \in O$. Let $I(r) = R(r,o,f)$ be the reward gained by visiting each node in $r$. Let the cost of traversal be given by $C(r) = C(v_s, v_i^r)$, where $v_i$ is the $i^{th}$ element in $r$, $\forall i \in [1,|r|]$. Let $L \in {\rm I\!R}_0^+ \cup \{\infty\}$ be the constraint limit. Using above notations, the hard-constraint problem can then be defined by equation \ref{eq_3}.
\begin{equation} \label{eq_3}
    \argmax_{r \subseteq V}I(r) \:\textit{subject to}\:C(r) \leq B \leq L
\end{equation}

We can relax the above hard-constraint by introducing a hyper-parameter $\delta$ to formulate a soft-constraint problem as shown in equation \ref{eq_4}.
\begin{equation} \label{eq_4}
    \argmax_{r \subseteq V}I(r) \:\textit{subject to}\: C(r) \leq B+\delta \leq L
\end{equation}
where $\delta \in {\rm I\!R}_0^+ \cup \{\infty\}$.

Informally, the solution to our stated problem is a set of ordered $k$-optimal goal nodes, such that the reward obtained by visiting the node is maximized while the path cost stays within a specified travelling budget $B$.

\section{Non-randomized Anytime Orienteering}
In this section, we propose a novel way of solving the problem formulation shown in equation \ref{eq_4} which is inspired by the general randomized algorithm for IPP problems \cite{IPPArora}.

The algorithm starts with a priority queue and creates $r$ subset s.t. $r \subseteq V$ \textbackslash\:$v_s$. Then, for each node in $r$, path cost $C(r)$ and node reward $I(r)$ is calculated. It is then ensured that the budget constraint is satisfied and the selected node is pushed into the priority queue with negative reward as the priority. We can pop the queue item with minimum priority $k$-times to find the $k$-most optimal goal nodes. This process is described in Algorithm \ref{Alg1}.

\begin{algorithm}
\SetAlgoLined
 \textbf{Inputs:} $G_s = (V, E), v_s, B, L, k, \delta$, $o \in O$, $f \subseteq F$ \\
 \textbf{Output:} $v_g = \{v_{g1},..,v_{gk}\}$ s.t. $v_{g1}>...> v_{gk}$ \\
 queue := new priority queue \\
 $v_g = \emptyset$ \\
 $r := r \subseteq V$ \textbackslash\:$v_s$ \\
 \For{$v_i\:\:in\:\:r$}{
   $I(r) = R(v_i,o,f)$        \:\:\:\:\texttt{//node reward} \\
   $C(r) = C(v_s, v_i)$     \:\:\:\:\texttt{//path cost} \\
  \If{$C(r)\leq B + \delta \leq L$}{
        $priority = -1 * I(r)$ \\
        queue.insert($v_i, priority$) \\
    }
 }
 \While{not queue.empty()}{
    $\rho$ := queue.pop-min()  \:\:\:\:\texttt{//best node} \\
    \If{$len(v_g) < k$}{
        $v_g:= v_g \cup \rho$
    }
 }
 \caption{Non-Randomized Anytime Orienteering to find k-optimal goals for a specialized graph}
 \label{Alg1}
\end{algorithm}

% \subsection{Properties \& Limitations}
% In this section, we will list some of the properties and drawbacks of the algorithm.

\subsubsection{Time Complexity.} If we assume a standard binary heap implementation of the priority queue, then the insertion and deletion time complexity is $O(log\:n)$, where $n$ is the size of the input \cite{heap1}. This can be further optimized by several customizations \cite{heap2}. Hence, the time complexity of our proposed algorithm for the best and the worst case can be stated as $ O(n-1\:*\:log\:n) + O(k\:*\:log\:n) \leq O(n\: log\:n)$.

\subsubsection{Space Complexity.} If we again assume a heap data structure implementation of the priority queue, then the space complexity of storing $n$ elements in the priority queue is $O(n)$ \cite{heap1}. Hence, the best and worst case space complexity of our proposed algorithm is $O(n)$.

\subsubsection{Limitations.} Our algorithm relies on the assumption that the graph is a specialized weighted directed graph with one central node (0 in-degree and $n$ out-degree) and $n$ isolated nodes connected with only one central node. Due to this assumption, the algorithm is efficient and applicable only for such versions of the specialized graph and cannot be extended implicitly to any general weighted directed graph.

\section{Experimental Results}
\begin{figure}[t]
\centering
\includegraphics[width=0.7\columnwidth]{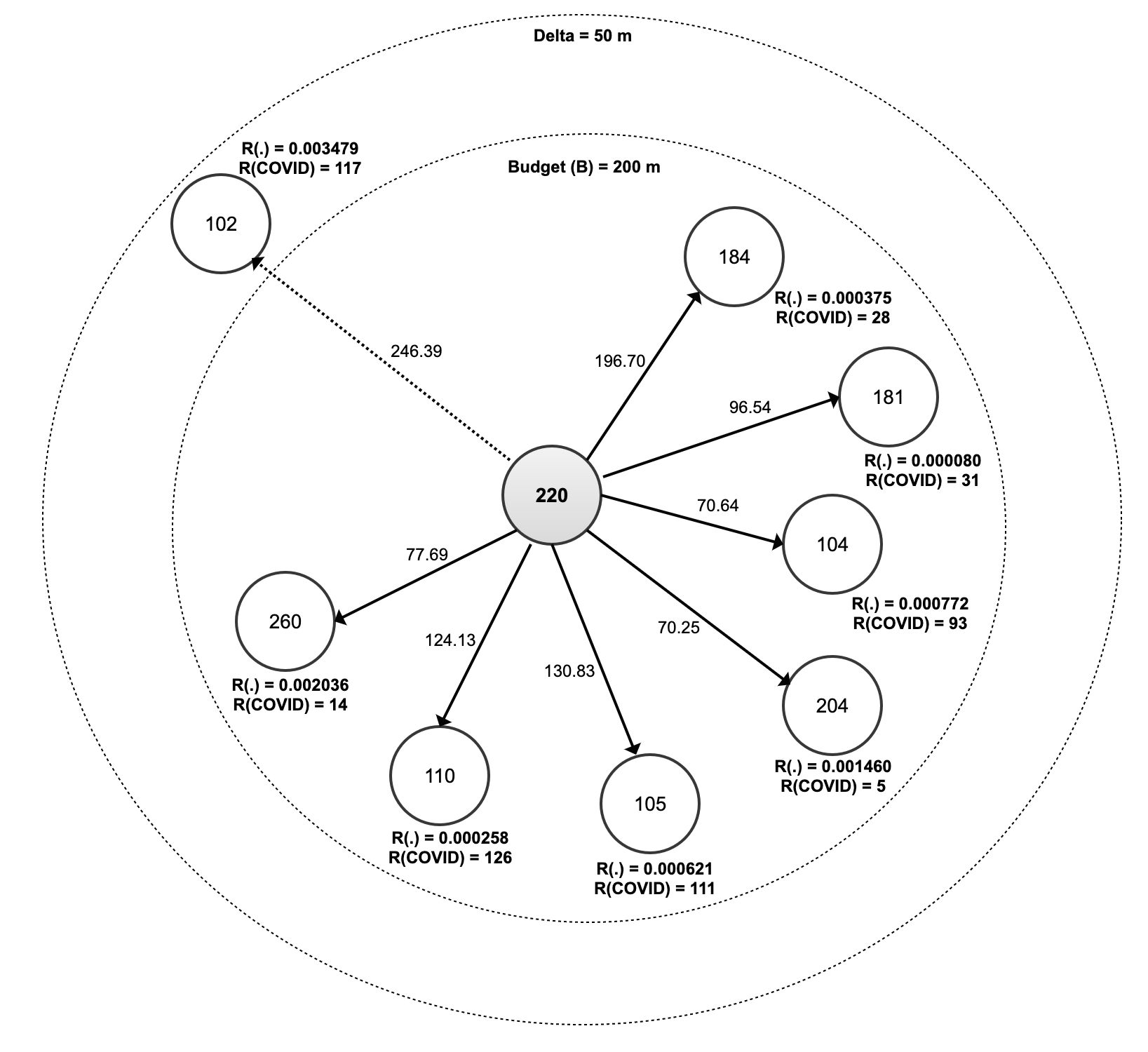} % Reduce the figure size so that it is slightly narrower than the column. Don't use precise values for figure width.This setup will avoid overfull boxes.
\caption{Finding $k=3$ most optimal nearest building from $v_s = 220$ that maximises the chances (reward) of booking a meeting room within $B=200$ meters and $\delta=50$ meters}
\label{fig1}
\end{figure}

In this section, we show experimental results for a domain-specific orienteering problem solved using our proposed algorithm. Here, our goal is to find the k-most optimal nearest building inside the Parkville campus of the University of Melbourne. These buildings should be within a specific radius ($B$) that maximises the chances (reward) of either booking a meeting room or using a toilet facility based on supply, demand and other preferences or factors. A specific scenario is shown in Figure \ref{fig1} where $R(.)$ are the rewards given by the buildings with no factors and $R(\textit{COVID})$ are the rewards based on COVID-19 lock-down restrictions.

Table \ref{tab1} shows the results for the stated scenario for 3-optimal nearest buildings using our proposed algorithm. In addition,  we were also able to simulate a COVID-19 restriction scenario by enhancing the reward function $R(r,o,f)$, obtaining results as shown in the Table \ref{tab2}.
\begin{table}[H] 
\centering
\resizebox{\columnwidth}{!}{%
\begin{tabular}{lrr}
\hline
\textbf{Goals} & \textbf{Cost}  & $R(.)$ \\ \hline
$v_{g1} = 260$             & 77.69   & 0.0020              \\
$v_{g2} = 204$             & 70.25   & 0.0014             \\ 
$v_{g3} = 104$           & 70.64   & 0.0007              \\ \hline
\end{tabular}
\quad
\begin{tabular}{lrr}
\hline
\textbf{Goals} & \textbf{Cost}  & $R(.)$ \\ \hline
$v_{g1} = 102$             & 246.39   & 0.0034              \\ 
$v_{g2} = 260$             & 77.69   & 0.0020              \\
$v_{g3} = 204$             & 70.25   & 0.0014               \\ \hline
\end{tabular}%
}
\caption{$k=3$ most optimal nearest buildings without any factors with $B=200\:m$ hard-constraint (left) and $B+\delta = 250\:m$ soft-constraint (right)}
\label{tab1}
\end{table}

\begin{table}[H]
\centering
\resizebox{\columnwidth}{!}{%
\begin{tabular}{lrr}
\hline
\textbf{Goals} & \textbf{Cost}  & $R(\textit{COVID})$ \\ \hline
$v_{g1} = 110$             & 124.13   & 126             \\
$v_{g2} = 105$             & 130.83   & 111             \\ 
$v_{g3} = 104$           & 70.64   & 93              \\ \hline
\end{tabular}
\quad
\begin{tabular}{lrr}
\hline
\textbf{Goals} & \textbf{Cost}  & $R(\textit{COVID})$ \\ \hline
$v_{g1} = 110$             & 124.13   & 126             \\
$v_{g2} = 102$             & 246.39   & 117             \\ 
$v_{g3} = 105$             & 130.83   & 111             \\ \hline
\end{tabular}%
}
\caption{$k=3$ most optimal nearest buildings in COVID lockdown situation with $B=200\:m$ hard-constraint (left) and $B+\delta = 250\:m$ soft-constraint (right)}
\label{tab2}
\end{table}

% \section{Conclusion}
% In this paper, we formulated a domain-specific problem into a general specialized graph based orienteering problem with budget constraints. We also proposed non-randomized anytime orienteering algorithm which solves the formulated problem with $O(n\:log\:n)$ time complexity. Preliminary results were also presented for a specific scenario to highlight the efficiency and usability of the algorithm.

\bibliography{references}

\end{document}